\documentclass[letterpaper]{article} 
\usepackage{aaai23}  
\usepackage{times}  
\usepackage{helvet}  
\usepackage{courier}  
\usepackage[hyphens]{url}  
\usepackage{graphicx} 
\usepackage{natbib}  
\usepackage{caption} 
\usepackage{booktabs}
\usepackage{multicol}
\usepackage{multirow}
\usepackage{xcolor}
\usepackage{array}

\title{Prompting Neural Machine Translation with Translation Memories}
\author{
    Abudurexiti Reheman\textsuperscript{\rm 1},
    Tao Zhou\textsuperscript{\rm 1},
    Yingfeng Luo\textsuperscript{\rm 1},
    Di Yang\textsuperscript{\rm 2},
    Tong Xiao\textsuperscript{\rm 1,\rm 2},
    Jingbo Zhu\textsuperscript{\rm 1,\rm 2}\thanks{Corresponding author.}
}
\affiliations{
    \textsuperscript{\rm 1}School of Computer Science and Engineering, Northeastern University, Shenyang, China\\
    \textsuperscript{\rm 2}NiuTrans Research, Shenyang, China\\

    rexiti\_neu@outlook.com, zhoutao\_neu@outlook.com, luoyingfengmail@163.com,\\ yangdi@niutrans.com, \{xiaotong, zhujingbo\}@mail.neu.edu.cn
}

\begin{document}

\maketitle

\begin{abstract}
Improving machine translation (MT) systems with translation memories (TMs) is of great interest to practitioners in the MT community. However, previous approaches require either a significant update of the model architecture and/or additional training efforts to make the models well-behaved when TMs are taken as additional input. In this paper, we present a simple but effective method to introduce TMs into neural machine translation (NMT) systems. Specifically, we treat TMs as prompts to the NMT model at test time, but leave the training process unchanged. The result is a slight update of an existing NMT system, which can be implemented in a few hours by anyone who is familiar with NMT. Experimental results on several datasets demonstrate that our system significantly outperforms strong baselines.
\end{abstract}

\section{Introduction}

Integrating TM is one of the commonly used techniques to improve real-world MT systems. In TM-assisted MT systems, it is often assumed that there is a database in which high-quality bilingual sentence pairs are stored. When translating an input sentence, the most (or top-K) similar sentence pair, which is retrieved from TM, is used to optimize the translation. From the perspective of practical application, this approach is particularly useful for MT, especially when sentences are highly repetitive, such as in translating technical manuals, legal provisions, etc. Previous works show that translation quality can be significantly improved when a well-matched TM sentence pair is provided both in Statistical Machine Translation (SMT) \cite{ma2011consistent,wang2013integrating,li2014discriminative} and Neural Machine Translation (NMT) \cite{gu2018search,khandelwal2020nearest}.

However, there are two major problems with this type of work in real-world applications. First, it is difficult to find such a TM dataset in most cases, especially when users can not share their TM data with the public for some reason. Second, previous approaches often require model changes, including training the model with TM \cite{bulte2019neural,hossain2020simple,xu2020boosting}, changing the NMT model architecture for TM integration \cite{gu2018search,bapna2019non,xia2019graph,he2021fast}, and introducing additional modules \cite{zhang2018guiding,he2019word,khandelwal2020nearest}. In this case, it is difficult to incorporate TM into the NMT system even if TM data is provided, since TM incorporation can not be accomplished on a generic decoder, and a deeply customized decoder is needed.

\begin{figure}[t]
    \centering
    \includegraphics[width=\columnwidth]{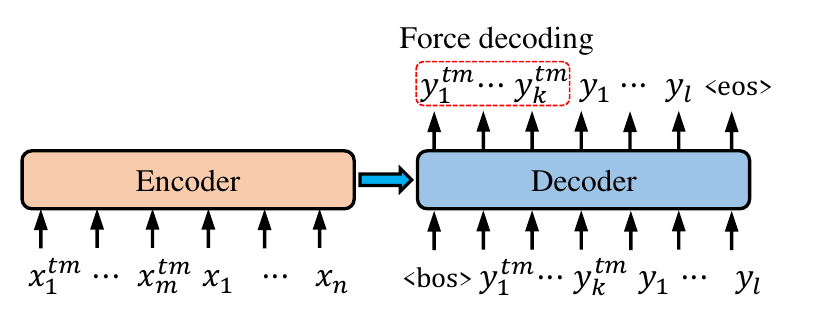}
    \caption{Structure of the proposed method. The source and target sentences of TM are concatenated with the input sentence and hypothesis in a specific concatenation template, respectively. The tokens in the target TM together with the concatenation template are generated in a forced manner. The lengths of the source and target TM together with the concatenation templates are $m$ and $k$, and the lengths of the input sentence and hypothesis are $n$ and $l$, respectively.}
    \label{fig 1}
\end{figure}

 Here, we address this problem by using few-shot learning \cite{wang2020generalizing}, which enables the system to quickly adapt to a small number of samples. The recent prevalence of prompt-based approaches \cite{brown2020language}, which transfer the original task into a generation task by designing an appropriate template without modifying the language model, gives us some inspiration that the retrieved TM can prompt the translation of the input sentence without modifying the NMT model. 

Based on this idea, we propose a simple approach to quickly adapt the NMT model in the few-shot TM scenario. Specifically, we treat TMs as prompts to the NMT model during the decoding process, with very small changes to the decoder. Our method can cover the advantages of conventional TM augmented methods and bring some new ideas, such as incorporating users' local historical translation data into NMT.

In order to prompt the translation with the retrieved TM, we design several templates to concatenate the source TM with the input sentence and feed the concatenated sentence into the model encoder. On the decoder side, we generate the target TM and the concatenation template in a forced way first, then let the model generate the other parts automatically. Regarding TM granularity, our method works well on sentence-level and fragment-level TM by designing appropriate templates. Experimental results on several datasets show that our method can further improve the translation quality on strong NMT models, and with comparable performance with the state-of-the-art.

\section{Background}

\subsection{NMT Decoding}
Suppose $x=\{x_1, ...,x_n\}$ is the source sentence, and NMT translates it into the corresponding target sentence $y=\{y_1, ...,y_m\}$ by using a trained NMT model. In practice, it turns the decoding into a searching problem, and a beam searcher is adopted to get the target sentence with the highest generation probability. Generally, an NMT model generates in an auto-regressive way. Therefore, the generation of each token relies on the source sentence and the generated prefix of the target sentence. The generation of the whole target sentence can be formulated as a conditional probability $P(y|x)$ described below:
\begin{eqnarray}
P(y|x)=\prod_{i=0}^{m}P(y_i|x,y_{<i}) \label{eq 1}
\end{eqnarray}
where $y_{<i}=\{y_1,y_2,...,y_{i-1}\}$ denotes the generated prefix tokens of target sentence at time-step $i$.

\subsection{TM}
TM is a database of language pairs that stores segments (such as fragments, sentences, paragraphs, or other sentence-like units) that have previously been translated by human translators for later use. It can provide identical or similar segments to help translate the input sentence. In the early stage, TM was widely used in Computer Aided Translation (CAT) \cite{Sarah2006Translators}. When translating an input sentence, the target sentence of TM is returned as the answer when an identical source sentence is found. In most cases, the translation result is obtained by fixing the most similar or top-K similar sentences retrieved from TM, since it is difficult to find a completely identical sentence. In the MT environment, TM is playing the same role as in CAT. Many approaches have been proposed to incorporate similar TM into MT systems to get a more accurate translation.

\section{Methodology}

In this section, we introduce our proposed method in detail from the perspectives of incorporating TM into NMT decoding, designing templates for concatenation, and retrieving similar TM.

\subsection{Incorporating TMs into NMT Decoding}

We incorporate TMs into the decoding process. For an input sentence $x$ and a retrieved TM sentence pair $\langle x^{tm},y^{tm}\rangle$, we concatenate $x^{tm}$ and $x$ in a specific concatenation template and feed it into the model encoder. On the decoder side, we first force the model to generate the exact tokens in $y^{tm}$ together with the concatenation template. Then the rest of the generation, which is returned as the answer, is done automatically without interfering. In practice, we set the generation probability of the tokens in the target TM and the concatenation template as 1 during the force decoding phase.

\subsection{Concatenation Templates}

\begin{table*}[]
\small
\centering
    \begin{tabular}{p{1.5cm}|p{1.1cm}|p{13.4cm}}
    \toprule
    \multicolumn{2}{l}{Input Sentence} & She gave us a full account of the traffic accident . \\
    \multicolumn{2}{l}{Source TM} & She gave the police a full account of the incident . \\
    \multicolumn{2}{l}{Target TM} & Sie gab der Polizei einen voll@@ ständigen Bericht über den Vorfall . \\
    \midrule
    \multicolumn{3}{c}{Sentence Level TMs}\\
    \midrule
    \multirow{2}*{Directly} & En/input & She gave the police a full account of the incident \textbf{\color{red}{.}} She gave us a full account of the traffic accident .\\
                            & De/input & \textless{bos}\textgreater{ } Sie gab der Polizei einen voll@@ ständigen Bericht über den Vorfall \textbf{\color{red}{.}} \{Hypothesis\} \\
    \cmidrule{2-3}
    \multirow{2}*{Comma} & En/input  & She gave the police a full account of the incident \textbf{\color{red}{,}} She gave us a full account of the traffic accident .\\
                        & De/input & \textless{bos}\textgreater{ } Sie gab der Polizei einen voll@@ ständigen Bericht über den Vorfall \textbf{\color{red}{,}} \{Hypothesis\} \\
    \cmidrule{2-3}
    \multirow{2}*{Semicolon}    & En/input & She gave the police a full account of the incident \textbf{\color{red}{;}} She gave us a full account of the traffic accident . \\
                                & De/input & \textless{bos}\textgreater{ } Sie gab der Polizei einen voll@@ ständigen Bericht über den Vorfall \textbf{\color{red}{;}} \{Hypothesis\} \\
    \cmidrule{2-3}
    \multirow{2}*{Conjunction}  & En/input & She gave the police a full account of the incident \textbf{\color{red}{. And ,}} She gave us a full account of the traffic accident . \\
                                & De/input & \textless{bos}\textgreater{ } Sie gab der Polizei einen voll@@ ständigen Bericht über den Vorfall \textbf{\color{red}{. Und ,}} \{Hypothesis\} \\
    \cmidrule{2-3}
    \multirow{2}*{Parenthesis}  & En/input & \textbf{\color{red}{(}} She gave the police a full account of the incident . \textbf{\color{red}{)}} She gave us a full account of the traffic accident . \\
                                & De/input & \textless{bos}\textgreater{ } \textbf{\color{red}{(}} Sie gab der Polizei einen voll@@ ständigen Bericht über den Vorfall . \textbf{\color{red}{)}} \{Hypothesis\} \\
    \midrule
    \multicolumn{3}{c}{Fragment Level TMs}\\
    \midrule
    \multirow{2}*{Parenthesis}  & En/input & \textbf{\color{red}{(}} She gave \textbf{\color{red}{) (}} a full account of the \textbf{\color{red}{)}} She gave us a full account of the traffic accident . \\
                                & De/input & \textless{bos}\textgreater{ } \textbf{\color{red}{(}} Sie gab \textbf{\color{red}{) (}} einen voll@@ ständigen Bericht über den \textbf{\color{red}{)}} \{Hypothesis\} \\
    
    \bottomrule
    \end{tabular}
\caption{An example of model input in our proposed method. Here, Directly, Comma, Semicolon, Conjunction, and Parenthesis denote our designed templates for concatenation, and En/input and De/input denote the input of the encoder and the decoder, respectively. The \textless{bos}\textgreater{ } token denotes the begin-of-sentence tag, and \{Hypothesis\} denotes the automatically generated part of the target sentence. \textbf{\color{red}{Red}} tokens denote the concatenation templates.}
\label{tab 1}
\end{table*}

\begin{figure*}[h]
    \centering
    \includegraphics[width=0.9\textwidth]{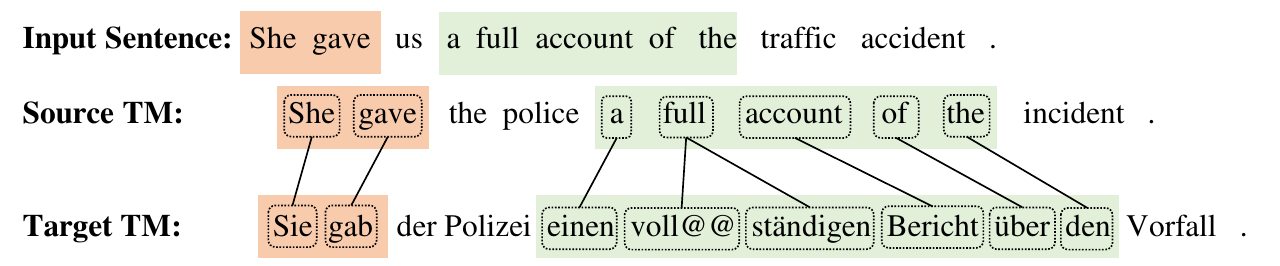} 
    \caption{An example of obtaining fragments for fragment-level TMs. For a given bilingual TM, common fragments between the input sentence and the source TM are acquired first, then the words in the target TM that align with the words in the common fragments are extracted. Common fragments and their corresponding fragments in Target TM are tagged by the same color box, and the lines denote word alignments.}
    \label{fig 2}
\end{figure*}

As our method incorporates TM on a pre-trained NMT model with no modifications, we must adhere to the following principles when designing the concatenation template: tokens in the concatenation template must be recognized by the NMT model and each part of the concatenated sentence maintain relatively complete semantics. Following this idea, we designed several templates for TMs in different granularity. In concatenation, the TM comes first on both the source and target sides, and a specific template is used on both sides. 

\paragraph{Sentence-level TM.} We concatenate sentence-level TMs in five different templates. As the example shown in Table \ref{tab 1}, our designed templates for sentence-level TMs are as follows:

(a) Concatenate directly. We concatenate them directly and add a period at the end of TMs if it is not ended up with punctuation marks.

(b) Concatenate with comma. Before concatenation, we replace the punctuation mark at the end of TMs with a comma or add a comma there if it is not ended up with any punctuation mark.

(c) Concatenate with semicolon. The concatenation is the same as the comma concatenation process, using a semicolon instead of the comma.

(d) Concatenate with conjunctions. In this template, we adopt conjunctions that express juxtaposed semantics, such as ``and" in English and ``und" in German. Specifically, we add a period at the end of the source and target TM if they are not ended up with any punctuation mark, then add a conjunction word in the corresponding language and a comma after that. Then we concatenate them with the input sentence and the hypothesis on the source and target side, respectively.

(e) Enclose in parentheses. We enclose both source and target TMs in parentheses, then perform the concatenation.

\paragraph{Fragment-level TM.} Unlike the sentence-level TM, we do some preprocessing on fragment-level TMs. First, we obtain the common fragments between the input sentence and source TM. Then, the tokens which are the translation of the words in the above common fragments are acquired from the target TM, using word alignment tools. After that, we construct the fragment-level TM using the tokens from the source and target TM above. An example of common fragments between input sentence and source TM and the word alignment between the source and target TM is given in Figure \ref{fig 2}, and its corresponding fragment-level TM is given in Table \ref{tab 1}. 

Specifically, for a given input sentence $x$ and a retrieved bilingual TM $\langle x^{tm},y^{tm}\rangle$, we acquire the encoder and decoder input for the NMT model in the following steps:

(a) Perform the longest common subsequence matching algorithm to $x$ and $x^{tm}$, and obtain the longest common subsequence $P_s = \{w_1,w_2,...,w_m\}$.

(b) Use word alignment tools to $x^{tm}$ and $y^{tm}$, and get the aligned subsequence $P_t = \{w^\prime_1,w^\prime_2,...,w^\prime_n\}$, which is corresponding to $P_s$, from $y^{tm}$.

(c) Group the words, which appear continuously in $x^{tm}$, from $P_s$ in their original order to form source TM fragments $P^\prime_s = \{f_1,f_2,...,f_i\}$.

(d) Group the words, which appear continuously in $y^{tm}$, from $P_t$ in the order of the correspondence with $P^\prime_s$ to form target TM fragments $P^\prime_t = \{f^\prime_1,f^\prime_2,...,f^\prime_j\}$.

(e) Concatenate each fragment in $P^\prime_s$ and $P^\prime_t$ with a specific template to form fragment-level TM, and directly concatenate them with input sentence and hypothesis as is done in sentence-level TM.

As for the concatenation template, we enclose each fragment in parenthesis to maintain its semantic integrity. In practice, we remove the fragments that consist of a single stop word and remove the punctuation at two sides of a fragment.

\subsection{Retrieving Similar TMs}

To retrieve the most similar bilingual TM for the input sentence, we use a word-level fuzzy matching strategy and remove punctuations and numbers from the sentence. Instead of retrieving from the whole TM database, we first employ the search engine library Apache Lucene \cite{bialecki2012apache} to retrieve the top 500 similar bilingual sentences from TM. Then we rerank them by adopting \emph{Fuzzy Match Score} (FMS) to obtain the most similar TM sentence pair. \emph{FMS} is a length normalized Levenshtein Distance \cite{yujian2007normalized}, known as Edit Distance:
\begin{eqnarray}
FMS(x,x^{tm})=1-\frac{LD(x,x^{tm})}{\textrm{max}(|x|,|x^{tm}|)}
\end{eqnarray}
where $LD(\cdot,\cdot)$ denotes the word level Levenshtein Distance, and $|\cdot|$ denotes word level length of a sentence.

\section{Experiments}

In order to verify the validity of our proposed method, we conducted several experiments on TM specialized translation task and domain adaptation task, respectively. We also put our approach into practice on a commercial NMT system to assess its usability in the practical setting. In the end, we investigated the impact of the NMT model, TM similarity, and input sentence length on translation quality.

\subsection{Datasets and Models}

For TM specialized translation tasks, we evaluated our method on two datasets: 1) DGT-TM, the entire body of European legislation in 22 European languages, on German-English in both directions (En-De and De-En) and 2) United Nations Parallel Corpus (UNPC), consisting of United Nations General Assembly Resolutions with translations in the six official languages, on English-Chinese (En-Zh), Russian-Chinese (Ru-Zh) and French-Chinese (Fr-Zh). These two datasets are relatively easy to retrieve TM sentences with a high degree of similarity. For the test set and TM database, we cleaned the above corpora first, then randomly selected 3,000 sentence pairs for the test dataset, whereas the remaining corpora were utilized as the TM database. For tokenization, we used NiuTrans \cite{xiao2012niutrans} word segmentation tool for Chinese and
Moses toolkit \cite{philipp2007moses} for other languages.

In addition, we performed an experiment using a homemade English-Chinese dataset (denoted as H-m in Table \ref{tab 2}) of 3401 sentences. Each test sentence has one bilingual TM sentence whose source side is similar to the test sentence. The statistics of these TM databases and the TM similarity ratios of retrieved TMs in FMS metric are shown in Table \ref{tab 2}.
 
In the domain adaptation task, following \citet{khandelwal2020nearest}, we used the multi-domain datasets from \citet{aharoni2020unsupervised}, which contains German-English bilingual datasets in five different domains: Medical, Law, IT, Koran, and Subtitles, respectively. We treated the training data in each domain as our TM database.

After cleaning the above corpora and splitting them into a test set and TM database, we retrieved the most similar TM for each test sentence from the TM database in an offline way and applied BPE \cite{2016Neural} to the test sets and TM with the BPE-codes provided by the pre-trained NMT models. To obtain the alignment information for the tokens in the TM source and target sentence, we trained a word aligner -- Mask-Align -- as proposed in \citet{chen2021maskalign}, and constructed corresponding fragment level TMs. The TM data scale and the TM similarity ratios of retrieved TMs in FMS metric are given in Table \ref{tab 3}.

\begin{table}[ht]
    \small
    \centering
    \begin{tabular}{p{0.8cm}|p{0.9cm}|p{0.7cm}<{\centering}|p{0.5cm}p{0.5cm}p{0.5cm}p{0.5cm}p{0.5cm}}
    \toprule
    \multirow{2}*{Corpus} & \multirow{2}*{Lang} & \multirow{2}*{\shortstack{TM\\scale}} & \multicolumn{5}{c}{TM FMS ratio} \\ 
              &  &  & [0, 0.2) & [0.2, 0.4) & [0.4, 0.6) & [0.6, 0.8) & [0.8, 1.0) \\ 
    \midrule
    \multirow{2}*{\shortstack{DGT\\-TM}} & En-De &3.1M  & 2\%  & 24\% & 16\% & 16\% & 42\% \\ 
        & De-En &3.1M  & 4\%  & 26\% & 17\% & 17\% & 36\% \\ 
    \midrule
    \multirow{3}*{UNPC} & En-Zh & 11.7M  & 2\%  & 44\% & 22\% & 11\% & 22\% \\ 
        & Fr-Zh & 11.5M  & 2\%  & 45\% & 18\% & 11\% & 23\% \\ 
        & Ru-Zh & 11.2M  & 8\%  & 46\% & 16\% & 9\% & 20\% \\ 
    \midrule
    H-m & En-Zh & -  & 18\%  & 30\% & 35\% & 23\% & 7\% \\ 
    \bottomrule
    \end{tabular}
    \caption{Sentence numbers in the TM databases and the similarity ratios of the retrieved TM.}
    \label{tab 2}
\end{table}
\begin{table}[h]
    \small
    \centering
    \begin{tabular}{p{1.1cm}|p{0.8cm}|p{0.5cm}p{0.5cm}p{0.5cm}p{0.5cm}p{0.5cm}}
    \toprule
    \multirow{2}*{Domain} &\multirow{2}*{\shortstack{TM\\scale}} & \multicolumn{5}{c}{TM FMS ratio} \\ 
               &    &    [0, 0.2) & [0.2, 0.4) & [0.4, 0.6) & [0.6, 0.8) & [0.8, 1.0) \\ 
    \midrule
    Medical  &248K  & 7\%  & 23\% & 20\% & 17\% & 33\% \\ 
    Law & 467K  & 8\%  & 31\% & 18\% & 14\% & 28\% \\ 
    IT & 223K  & 14\%  & 18\% & 28\% & 26\% & 14\% \\
    Koran & 18K & 2\%  & 26\% & 33\% & 28\% & 11\% \\
    Subtitles & 500K  & 3\%  & 27\% & 43\% & 23\% & 4\% \\
    \bottomrule
    \end{tabular}
    \caption{Sentence numbers in the TM database in each domain and the similarity ratios of the retrieved TM.}
    \label{tab 3}
\end{table}

As for the pre-trained NMT model, we applied Facebook's WMT19 De-En, En-De model \cite{Nathan2019Facebook} and NiuTrans' WMT20 En-Zh model \cite{Yuhao2020TheNiuTrans} as our base model. All of these models are very competitive that they trained on more than 20 million training data and 10 million extra back-translated data.

\subsection{Main Experiment}
Our main experiment involves the TM specialized translation task, the domain adaptation task, and the implementation on commercial NMT system.

\paragraph{TM Specialized translation.}

In this experiment, we decoded the DGT-TM En-De and De-En test sets using facebook's WMT19 En-De and De-En models \cite{Nathan2019Facebook}, respectively. Besides, we decoded the UNPC En-Zh test set and the homemade En-Zh test set with NiuTrans' WMT20 En-Zh model \cite{Yuhao2020TheNiuTrans}. For the Mask-Align model training, we used WMT20 En-Zh training data for the En-Zh aligner and DGT-TM's TM database for En-De and De-En aligner.
From the experimental results in Table \ref{tab 4}, we have the following observations.

First, in sentence-level TM, the BLEU score on DGT-TM De-En, En-De, and homemade En-Zh test sets increased significantly, with maximum BLEU score increases of 8.63, 5.74, and 7.74 points, respectively. Meanwhile, the translation improved slightly on the UNPC En-Zh test set in directly, semicolon, and parenthesis concatenations. Second, in fragment-level TM, the BLEU score increased by about 1 to 2 points or even decreased, compared to the baseline (without TM). The main reason for this is that the NMT model is trained on sentence-level training data rather than sentence pieces. In addition, it is also affected by the performance of the word aligner, which may provide error alignment information.

\begin{table}[h]
    \small
    \centering
    \begin{tabular}{p{0.3cm}|p{0.9cm}|p{0.9cm}p{0.9cm}p{0.9cm}p{0.9cm}}
    \toprule
    \multicolumn{2}{c|}{Corpus} & \multicolumn{2}{c}{DGT-TM} & UNPC & H-m \\ 
    \multicolumn{2}{c|}{Lang}   & De-En & En-De & En-Zh & En-Zh \\ 
    \midrule
    \multicolumn{2}{c|}{W/o TM} & 45.40 &39.03 &41.42 & 46.43 \\
    \midrule
    \multirow{5}* {\rotatebox{90}{Sentence TM}} 
                & Directly & 53.74 & 44.32 & 41.70 & 52.97  \\
                & Comma & 52.44 & 43.03 & 41.33 &  51.85 \\
                & Semico & 53.42 & 44.54 & \textbf{42.31} &  52.89 \\
                & Conjunc & 53.65 & 44.00 & 41.15 &  \textbf{54.17} \\
                & Parenth & \textbf{54.03} & \textbf{44.77} & 41.90 &  53.87 \\
    \midrule
    \multicolumn{2}{c|}{Fragment TM} & 47.21 & 41.65 & 39.85 & 47.67 \\
    \bottomrule
    \end{tabular}
    \caption{Experimental results on the DGT-TM En-De, De-En, UNPC En-Zh, and the home-made En-Zh test sets.}
    \label{tab 4}
\end{table}

\paragraph{Domain Adaptation.}

Following the $k$NN-MT \cite{khandelwal2020nearest} and its optimized counterparts, we conducted the domain adaptation experiment and compared our method with $k$NN-MT. We applied Facebook’s WMT19 De-En model \cite{Nathan2019Facebook} for decoding.  Experimental results are given in Table \ref{tab 5}. 

From the table, we can find that our method improves the translation in all domains except Subtitles, with maximum BLEU score improvements of 2.54, 3.05, 4.64 in IT, Law, and Medical domains, respectively, whereas in Koran the result is only 0.42 BLEU score higher. The fragment-level TM method has the same tendency as the above experiment, which is slightly higher only in IT and Medical domain than the baseline. Besides, the improvement of our method is less than $k$NN-MT in every domain. The original design of our approach leads to this result. Our method retrieves the most similar single TM and leverages the knowledge it contains to improve the translation. How to leverage the knowledge provided by TM is fully dependent on the NMT model itself. While $k$NN-MT introduces an extra module to incorporate the information explicitly from multiple similar context vectors. 

The main advantage of our method over $k$NN-MT is that our method performs the retrieval based on string similarity, and there is no need to store the context vectors, which saves a lot of storage space. At the same time, $k$NN-MT searches the context vectors in each beam in every timestep, which is much slower than the vanilla NMT. However, our method searches the TM only once and generates two sentences (target TM and the hypothesis) in the way of a vanilla NMT. This will make our method much faster than $k$NN-MT.

\begin{table}[h]
    \small
    \centering
    \begin{tabular}{p{0.2cm}|p{0.7cm}|p{0.8cm}p{0.8cm}p{0.8cm}p{0.8cm}p{0.8cm}}
    \toprule
    \multicolumn{2}{c|}{Domains} & IT & Koran & Law & Medical & Subtitles \\ 
    \midrule
    \multicolumn{2}{c|}{$k$NN-MT} & 45.82 & 19.45 & 61.78 & 54.35  & 31.73\\
    \midrule
    \multicolumn{2}{c|}{W/o TM} & 38.09 &17.11 & 45.92 & 41.14 & \textbf{29.45}\\
    \midrule
    \multirow{5}* {\rotatebox{90}{Sentence TM}} 
                & Directly & 40.19 & 17.20 & 48.78 &  45.29 & 28.18\\
                & Comma & 39.20 & 16.46 & 47.42 &  43.47 & 25.09\\
                & Semico & 39.74 & 17.09 & 48.91 &  44.93 & 26.44\\
                & Conjunc & 40.13 & 17.03 & \textbf{48.97} &  45.13 & 27.68\\
                & Parenth & \textbf{40.63} & \textbf{17.53} & 48.31 &  \textbf{45.78} & 29.03\\
    \midrule
    \multicolumn{2}{c|}{Fragment TM} & 39.38 & 16.49 & 45.58 & 43.31 & 28.24  \\
    \bottomrule
    \end{tabular}
    \caption{Experimental results on multi-domain datasets.}
    \label{tab 5}
\end{table}

\paragraph{Implementation on Commercial NMT System.}
We implemented our proposed method on a commercial NMT system -- NiuTrans Enterprise -- to evaluate our method's applicability in the real-world environment. We experimented on UNPC En-Zh, Fr-Zh, and Ru-Zh, without modifying the NMT model, even not aware of what kind of NMT model is used. The word aligners for fragment-level TM on Fr-Zh and Ru-Zh are trained on Fr-Zh and Ru-Zh TM databases, respectively. Experimental results in Table \ref{tab 6} show that the maximum improvement of sentence-level TM on En-Zh, Fr-Zh, and Ru-Zh are 2.94, 3.55, 3.06 BLEU points, respectively, and the fragment-level TM approach still get lower BLEU scores than the baseline. The NMT models used in this experiment have been trained on much more high-quality training data than other models used in the above experiments. The experimental results demonstrate that even strong commercial NMT systems can be further improved when similar TMs provided and that the sentence-level TM approach can be applied in real-world situations where similar TMs for input sentences are available.

\begin{table}[h]
    \small
    \centering
    \begin{tabular}{p{0.3cm}|p{0.9cm}|p{0.9cm}p{0.9cm}p{0.9cm}}
    \toprule
    \multicolumn{2}{c|}{Lang}   & En-Zh & Fr-Zh & Ru-Zh \\ 
    \midrule
    \multicolumn{2}{c|}{W/o TM} &41.59  & 29.83 & 35.62 \\
    \midrule
    \multirow{5}* {\rotatebox{90}{Sentence TM}} 
                & Directly & 44.18 & 33.10 & 37.94 \\
                & Comma    & 43.85 & 31.63 & 37.36 \\
                & Semico  & 44.48 & \textbf{33.38} & 37.89 \\
                & Conjunc  & 44.16 & 33.04 &  37.97  \\
                & Parenth  & \textbf{44.53} &  33.05 & \textbf{38.68}  \\
    \midrule
    \multicolumn{2}{c|}{Fragment TM} & 38.74 & 27.78  &  32.65  \\
    \bottomrule
    \end{tabular}
    \caption{Experimental results on a commercial NMT system.}
    \label{tab 6}
\end{table}

\subsection{NMT Model's Effect on Translation}
In our proposed method, the generation of each token in the target sentence relies on source TM, input sentence, target TM and the generated part of target sentence, and the target TM is generated in a forced way. Therefore, the translation depends on the translation ability of the NMT model, ``strong'' models improve greater, and ``weak'' models improve less or even get worse results. In order to investigate to what extent the results depend on the NMT model's ``strength'', we conducted a series of experiments on the UNPC En-Zh test set (see Table \ref{tab 2}) with different NMT models. We measure the translation ability of a model in terms of the training data scale and the model architecture. So, we trained several NMT models using WMT20 En-Zh training data, from the perspectives of training data scale and model architecture.

\paragraph{Training Data Scale.}
The training data of WMT20 En-Zh has 20 million bilingual sentences. We uniformly split them into four parts after shuffling, then trained four NMT models with different data scales, in which the first model is trained on the first 5 million datasets, the second model is trained on the first and second 5 million datasets, and so on. All of the models are transformer big models proposed in \citet{NIPS2017_attention}. The experimental results are given in Table \ref{tab 7}.

\begin{table}[h]
    \small
    \centering
    \begin{tabular}{p{1.1cm}|p{0.7cm}p{0.7cm}p{0.7cm}p{0.7cm}p{0.7cm}p{0.7cm}}
    \toprule
    Models & b5M & b10M & b15M & b20M & ba20M & bb20M\\ 
    \midrule
    W/o TM  & 41.11 & 41.40 & 42.53 & 43.19 & 41.47 & 42.53    \\
    \midrule
                Directly & 41.77 & 42.45 & 44.27 & 44.26 & 41.87 & 43.75   \\
                Comma & 41.29 & 41.88 & 43.79 & 43.99 & 41.18  &  43.36  \\
                Semico & \textbf{42.27} & 42.45 & \textbf{44.64} & \textbf{45.13} & \textbf{42.12} &  \textbf{44.28}  \\
                Conjunc & 41.48 & 42.04 & 43.87 & 44.33 & 41.54 &  43.09  \\
                Parenth & 41.68 & \textbf{42.63} & 44.49 & 44.53 & 41.93 & 43.85  \\
    \midrule
    Max $\Delta$ & 1.16 & 1.23 & 2.11 & 1.94 & 0.65 & 1.75   \\
    \bottomrule
    \end{tabular}
    \caption{Experimental results on UNPC En-Zh test set with different NMT models, including four transformer big models trained on 5 million, 10 million, 15 million, and 20 million training data (denoted as b5M, b10M, b15M, b20M, respectively), and a transformer base and a bigger model trained on 20 million training data (denoted as ba20M and bb20M, respectively), max $\Delta$ denotes the maximum improvement comparing to decoding without TM.}
    \label{tab 7}
\end{table}

\paragraph{Model Architecture.}
In this experiment, we investigate the impact of the model architecture on our proposed method. We chose WMT20 En-Zh dataset with 20 million bilingual sentences in the above experiment and trained the transformer base, big and bigger models, respectively. Their attention heads, hidden sizes, and filter sizes are (8, 512, 2048), (16, 1024, 4096), and (24, 1536, 6144), respectively. From the experimental results in Table \ref{tab 7}, we have the observations below.

For the same model architecture, with the increase of training data scale, the translation ability of the model is getting stronger, and the BELU improvement of our method is also higher compared with the baseline, as the maximum BLEU score improvement of the models b15M and b20M are higher than that of b5M and b10M. In addition, for the models trained on the same training data, the ba20M model is ``weaker'' than the b20M and bb20M models, and the maximum BLEU score improvement is also lower than the latter two models. We can find a similar phenomenon if we look back to Table \ref{tab 4} and Table \ref{tab 6}. The dataset for UNPC En-Zh is the same in these two experiments, and the NMT model of the commercial NMT system is much ``strong'' than the NMT model used in table \ref{tab 4}, and the maximum improvement of the former is an 2.94 BLEU points, whereas the latter's is 0.89 BLEU points. From these results, we conclude that the sentence-level TM approach of our method can further improve strong baselines, and the ``stronger'' the NMT model, the greater the improvement will be.

\subsection{TM Similarity}

As our method obtain useful information from a single TM sentence pair, the translation result is influenced by the similarity of the retrieved TM. TMs with high similarity provide more useful information for the translation, while less similar ones may introduce noise to decoding. In this experiment, we explore the similarity threshold that a TM can help or not. We decoded the UNPC En-Zh test set using NiuTrans’ WMT20 En-Zh model \cite{Yuhao2020TheNiuTrans}. Specifically, the test set is divided into various portions based on the similarity score, and each portion of the test set is decoded individually both with the sentence-level TM approach and baseline (without TM). 

\begin{figure}
    \centering
    \includegraphics[width=\columnwidth]{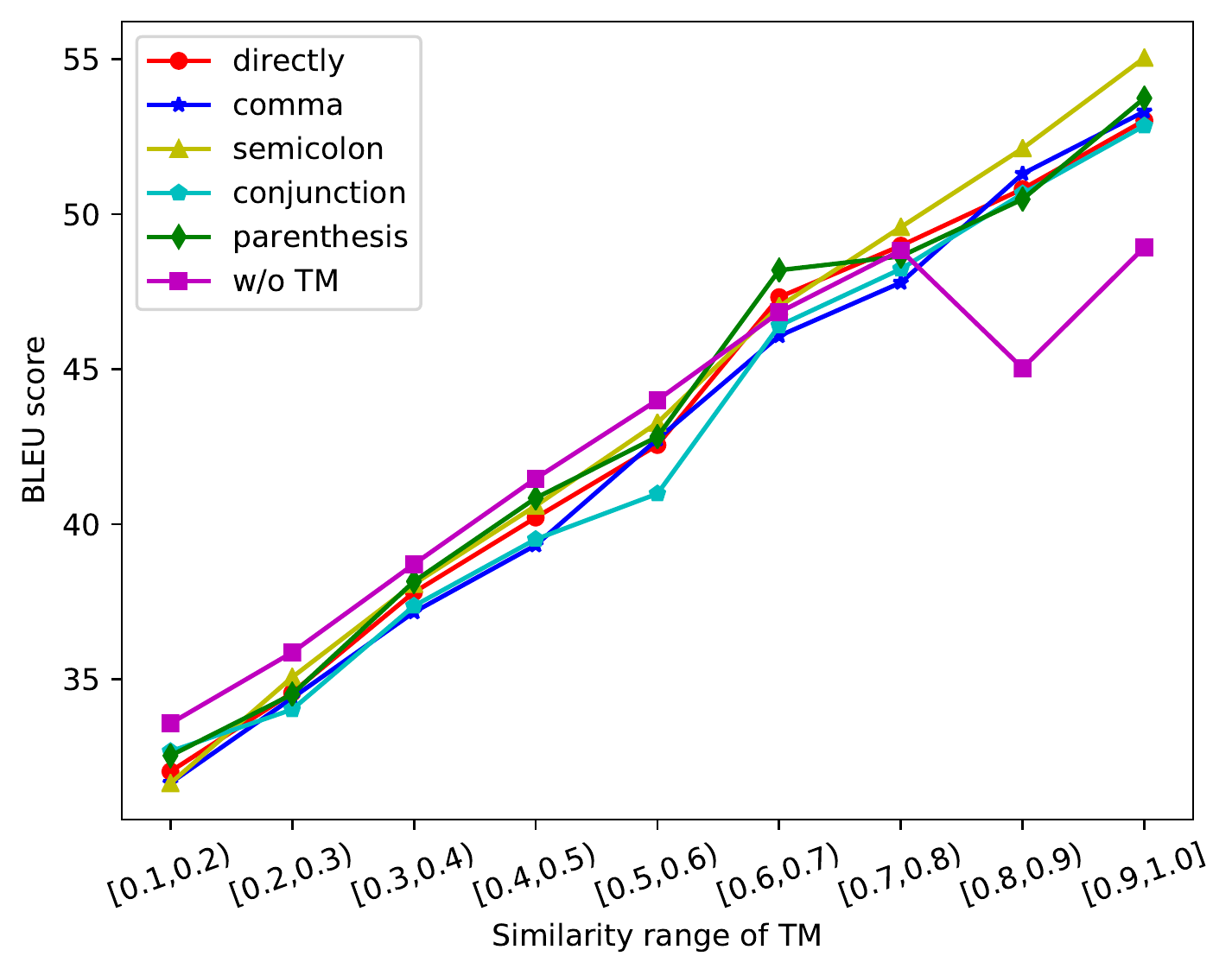}
    \caption{BLEU scores on different similarity ranges.}
    \label{fig 3}
\end{figure}

From the experimental results in Figure \ref{fig 3}, we can find that the BLEU scores of our method are lower than the baseline when the similarity score is lower than 0.6; when it is between 0.6 and 0.8, some of our concatenation methods perform better than the baseline with a marginal advantage; when it is higher than 0.8, all of the concatenation methods outperform the baseline significantly, with a maximum improvement of 7.09 and 6.11 BLEU points, respectively. Therefore, the FMS threshold for sentence-level TM of the NiuTrans WMT20 En-Zh model is 0.8. 

The threshold is determined by the ``strength'' of the NMT model that ``strong'' models are more robust to obtaining useful information and avoiding noises introduced by less similar TMs. Thus, ``strong'' models have lower FMS thresholds. In practice, the threshold can be used to decide whether to apply TM for decoding.

\subsection{Sentence Length}

\begin{figure}
    \centering
    \includegraphics[width=\columnwidth]{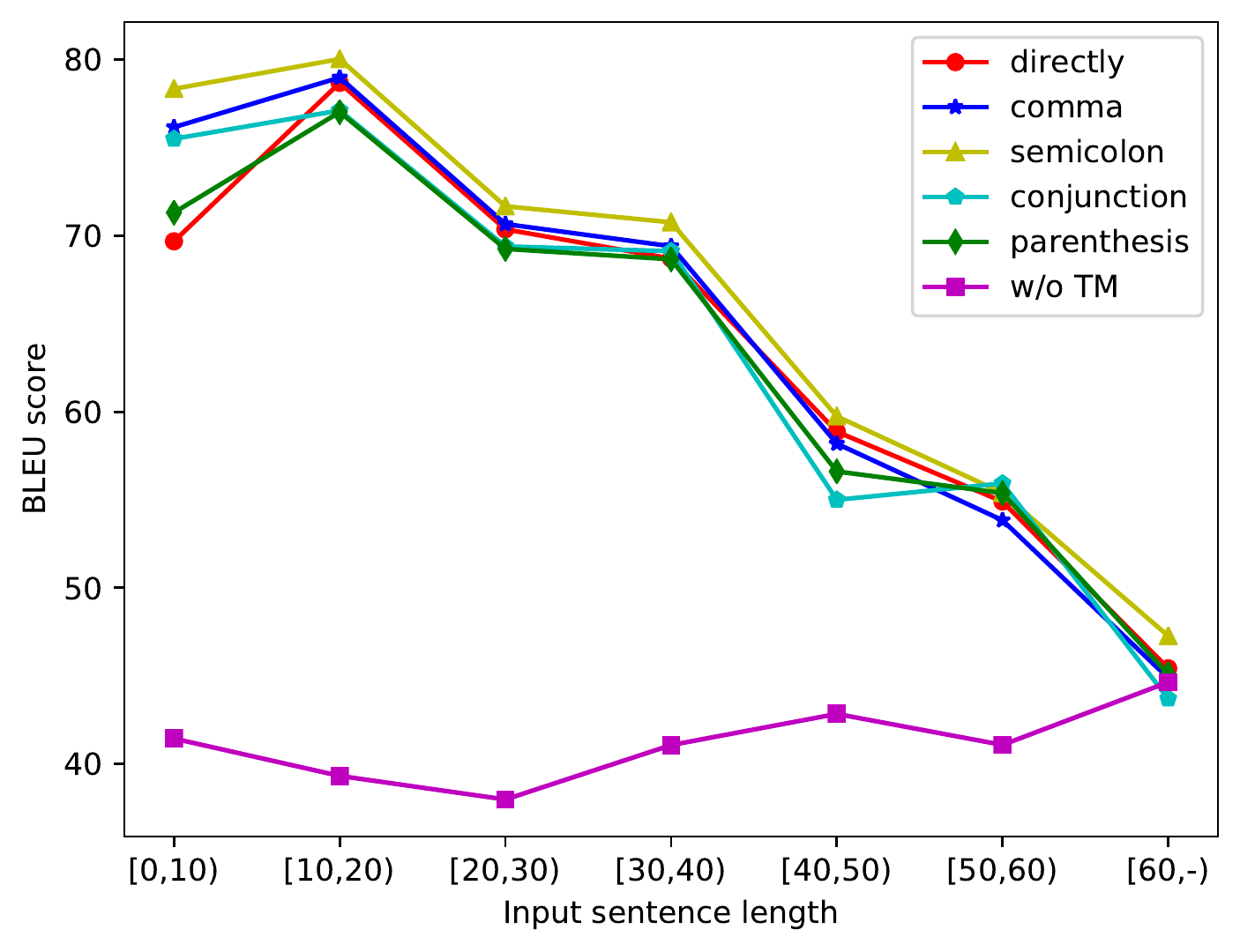}
    \caption{input sentence length impact on BLEU.}
    \label{fig 4}
\end{figure}

In this section, we investigate the impact of input sentence length on translation. To avoid the TM similarity influencing the experimental results, we used the test set itself as the retrieved TM, which means that the TMs are 100\% similar to the input sentences. We split the test set into groups according to the length of the input sentence, and uniformly chose 170 sentences from each group as the test set (the minimum sentence number of the original groups is 171). The experimental results are given in Figure \ref{fig 4}.

From the experimental results, we can find a sharp tendency that the performance of our proposed method decreases and eventually be comparable to the baseline as the sentence length increases. The performance of the baseline, however, tends to be steady. This is also determined by the initial design of our method. For an input sentence, we employ an NMT model that has been trained on a sentence-level dataset. As we concatenate the source TM with the input sentence before feeding them into the NMT model, the length of the input for the model encoder will be doubled. In this way, the whole sentence length of a lengthy sentence and its corresponding TM will deviate from the training model's sentence length distribution. This is the main cause of our method's performance in the figure degrading on lengthy sentences. Therefore, our method can not handle long sentences well even though a highly similar sentence is retrieved. After calculation, we find that the average sentence length of the UNPC En-Zh test set and homemade En-Zh test set in Table \ref{tab 4} are 29.58 and 10.99. This is why the latter can improve the translation more significantly than the former even though there are fewer sentences with high similarity than there are in the former.

\section{Related Work}

Many studies have been conducted in recent years to enhance MT quality using TMs. With the emergence of NMT, the MT community is seeing an increasing interest in TM research. There are mainly two research lines for TM integration into NMT: constraining the decoding process with TM and using TM to train a more powerful NMT model. 

The main idea of the first research line is to increase the generation probability of some target words based on TM. \citet{zhang2018guiding} constrained the decoding process by increasing the generation possibility of the target words which are in the aligned slices extracted from the TM. Following this work, \citet{he2019word} added positional information for words in the TM slices. \citet{li2018one} and \citet{farajian2017multi} embedded the retrieved TM information into the NMT model by fine-tuning the NMT model with TMs before translating the input sentence. Instead of incorporating text level TM, $k$NN-MT retrieved TMs from dense vectors \cite{khandelwal2020nearest}. First, they created a key-value datastore from the TM database, where the key is the translation context vector of each time step, and the value is the true target token. In the inference time, $k$NN-MT interpolates the generation probability of the NMT model and retrieved similar target distribution from that datastore at each time step. Following $k$NN-MT, several researches optimized $k$NN-MT from different perspectives. \citet{meng2021fast} accelerated the inference process by narrowing the search range, instead of searching from the entire data store. By introducing a lightweight meta-k network, \citet{zheng2021adaptive} dynamically determines how many neighbors should be introduced. \citet{wang2021faster} further accelerated $k$NN-MT inference by constraining search space when constructing the data store. Instead of retrieving a single token, \citet{Pedro2022Chunk-based} retrieved chunks of tokens from the data store to speed up $k$NN-MT.

The second research line aims to train the generation model to learn how to deal with the retrieved TMs. \citet{bulte2019neural,xu2020boosting,minh2020priming} used a data augmentation way to concatenate the retrieved TM with input sentence during training. One sub-method of \citet{minh2020priming} generated the target prefix in a forced way when decoding, while they need to train the NMT model with a similar concatenation pattern first. Some researches modified the NMT model architecture for better TM integration. \citet{cao2018encoding} and \citet{gu2018search} introduced a gating mechanism module to control the signal from the retrieved TM. \citet{qian2020Learning} designed an additional transformer encoder to integrate the target sentence of TM through the attention mechanism. In \citet{xia2019graph}, the retrieved multiple TMs are compressed into a graph structure for speed up and space savings and then are integrated into the model via the attention mechanism. \citet{he2021fast} proposed a lightweight method to incorporate the target sentence of retrieved TM in an extra attention module. Unlike all of the above methods, \citet{cai2021neural} proposed a method to incorporate monolingual TM into NMT, and the target sentence retriever and NMT model are trained jointly.

\section{Conclusion and Future Work}

In this paper, we propose a simple but effective method to incorporate TM into NMT decoding without modifying the pre-trained NMT model. Specifically, we treat the retrieved TMs as prompts for the translation of the input sentence by concatenating the source TM with the input sentence and generating the target token in a forced way. Experiments on the TM specialized translation task, domain adaptation task, and implementation on commercial MT system verify the effectiveness of our method. Our method is easy to implement and can be applied to customize a TM-incorporated machine translation system for TM data on the user side. Our method in this paper suffers from TM sentences with low similarity scores and long sentences. In the future, we will investigate more effective methods to alleviate the drawbacks of our methods in low similarity TM and long sentence translation situations.

\section{Acknowledgments}

This work was supported by National Key R\&D Program of China (No. 2020AAA0107904). We are very thankful to anonymous reviewers for their comments.

\bibliography{custom}

\end{document}